\documentclass[conference]{IEEEtran}

\usepackage{booktabs} 

\usepackage{blindtext, graphicx}
\usepackage{epsfig,xcolor}
\usepackage{multirow}
\usepackage{subfig}

\usepackage[noadjust]{cite}

\usepackage{amsmath,amssymb,amsfonts}
\usepackage{algorithmic}
\usepackage{graphicx}
\usepackage{xcolor,colortbl}

\newif\iffinal

\finaltrue

\iffinal
  \newcommand{\ian}[1]{}
  \newcommand{\ryan}[1]{}
  \newcommand{\kyle}[1]{}
  \newcommand{\zhuozhao}[1]{}
  \newcommand{\ben}[1]{}
  \newcommand{\logan}[1]{}
\else
  \newcommand{\ian}[1]{{\textcolor{red}{ Ian: #1 }}}
  \newcommand{\ryan}[1]{{\textcolor{magenta}{ Ryan: #1 }}}
  \newcommand{\kyle}[1]{{\textcolor{purple}{ Kyle: #1 }}}
  \newcommand{\zhuozhao}[1]{{\textcolor{blue}{ Zhuozhao: #1 }}}
  \newcommand{\ben}[1]{{\textcolor{olive}{ Ben: #1 }}}
  \newcommand{\logan}[1]{{\textcolor{teal}{ Logan: #1 }}}
\fi

\usepackage{url}
\usepackage{hyperref}
\usepackage{listings}
\usepackage{caption}
\usepackage{mathtools}

\definecolor{background}{HTML}{EEEEEE}
\definecolor{delim}{RGB}{20,105,176}
\colorlet{punct}{red!60!black}
\colorlet{numb}{magenta!60!black}

\lstdefinelanguage{json}{
    belowcaptionskip=1\baselineskip,
    basicstyle=\scriptsize\ttfamily,
    showstringspaces=false,
    breaklines=true,
    frame=tb,
		captionpos=t,
    literate=
     *{:}{{{\color{punct}{:}}}}{1}
      {,}{{{\color{punct}{,}}}}{1}
      {\{}{{{\color{delim}{\{}}}}{1}
      {\}}{{{\color{delim}{\}}}}}{1}
      {[}{{{\color{delim}{[}}}}{1}
      {]}{{{\color{delim}{]}}}}{1},
}

\begin{document}
\bstctlcite{IEEEexample:BSTcontrol}

\title{DLHub: Model and Data Serving for Science}

\author{\IEEEauthorblockN{Ryan Chard\IEEEauthorrefmark{1},
Zhuozhao Li\IEEEauthorrefmark{2},
Kyle Chard\IEEEauthorrefmark{3}\IEEEauthorrefmark{1},
Logan Ward\IEEEauthorrefmark{2}\IEEEauthorrefmark{1},
Yadu Babuji\IEEEauthorrefmark{3},
Anna Woodard\IEEEauthorrefmark{2},\\
Steven Tuecke\IEEEauthorrefmark{3},
Ben Blaiszik\IEEEauthorrefmark{3}\IEEEauthorrefmark{1},
Michael J. Franklin\IEEEauthorrefmark{2}, and
Ian Foster\IEEEauthorrefmark{1}\IEEEauthorrefmark{2}\IEEEauthorrefmark{3}, }
\IEEEauthorblockA{\IEEEauthorrefmark{1}Data Science and Learning Division, Argonne National 
Laboratory, Argonne, IL, USA\\}
\IEEEauthorblockA{\IEEEauthorrefmark{2}Department of Computer Science, University of Chicago, 
Chicago, IL, USA}
\IEEEauthorblockA{\IEEEauthorrefmark{3}Globus, University of Chicago, Chicago, IL, USA\\}
}

\maketitle
\begin{abstract}

While the Machine Learning (ML) landscape is evolving rapidly, 
there has been a relative lag in the development of the 
``learning systems''
needed to enable 
broad adoption. 
Furthermore, few such systems are designed to support the specialized
requirements of scientific ML. 
Here we present the Data and Learning Hub for science (DLHub), a 
multi-tenant system that provides both model repository
and serving capabilities with a focus on science applications. 
DLHub addresses two significant shortcomings in current systems. 
First, its self-service model repository
allows users to share, publish, verify, reproduce, and reuse models,
and addresses concerns related to model reproducibility by packaging
and distributing models and all constituent components. 
Second, it implements scalable and low-latency 
serving capabilities that can leverage parallel
and distributed computing resources to democratize
access to published models through a simple web interface. 
Unlike other model serving frameworks, DLHub 
can store and serve any Python~3-compatible model or processing function, plus multiple-function
pipelines.
We show that relative to other
model serving systems including TensorFlow Serving, SageMaker, and Clipper,
DLHub provides greater capabilities, comparable performance without memoization
and batching, and significantly better performance when the latter two techniques can be employed.
We also describe early uses of DLHub for scientific applications.
\end{abstract}



\section{Introduction}
\label{sec:introduction}

\begin{figure}[h]
 \centering
 \includegraphics[width=\columnwidth,trim=0.2in 4in 3.6in 0.4in,clip]{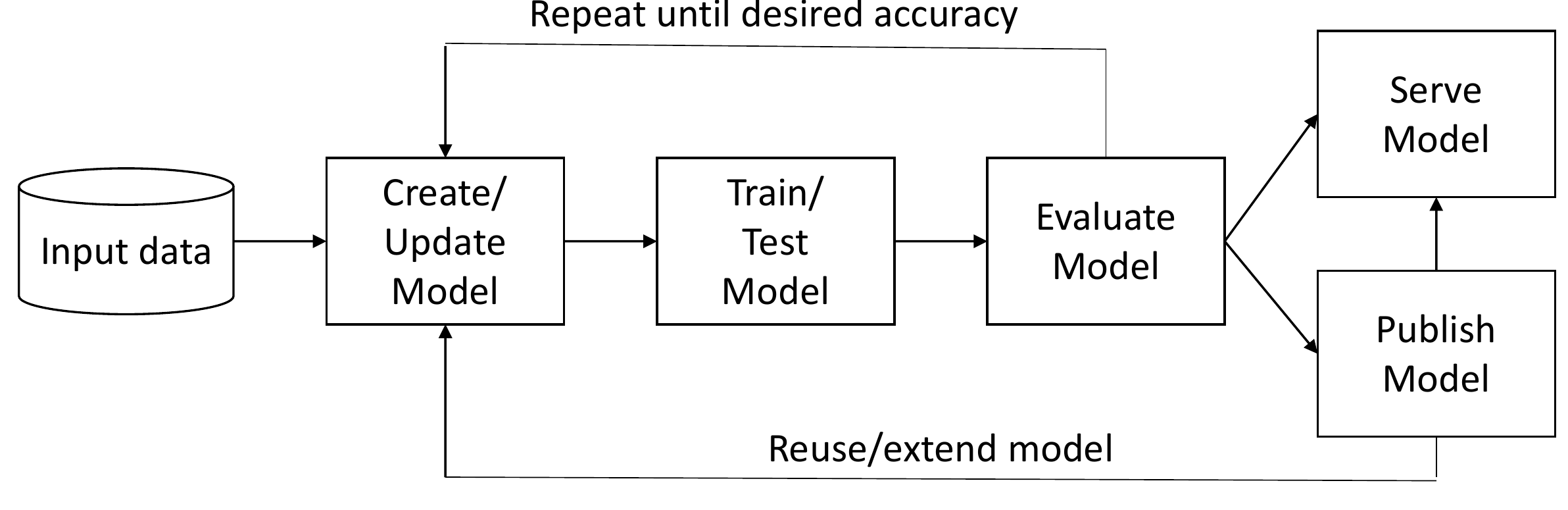}
 \caption{ML lifecycle, adapted from Miao et al.~\cite{miao2017towards}}
 \label{fig:lifecycle}
\end{figure}

Machine Learning (ML) is disrupting nearly every 
aspect of computing. Researchers now turn to ML methods 
to uncover patterns in vast data collections and to make 
decisions with little or no human input.
As ML becomes increasingly pervasive, new systems are required
to support the development, adoption, and application of ML. 
We refer to the broad class of systems designed to support
ML as ``learning systems.''
Learning systems need to support the entire ML lifecycle (see \figurename~\ref{fig:lifecycle}), 
including model development~\cite{balaprakash2016automomml,GoogleAutoML}; 
scalable training across potentially tens of thousands of
cores and GPUs~\cite{sagemaker}; model publication and sharing~\cite{avsec2018kipoi};
and low latency and high-throughput inference~\cite{crankshaw2017clipper}; 
all while encouraging best-practice software engineering when developing
models~\cite{miao2017towards}.

While many systems focus on building and training ML
models~\cite{jia2014caffe,abadi2016tensorflow,sagemaker},
there is a growing need for systems that support other
stages of the ML lifecycle and in particular those important 
in science, such as publishing models used 
in the literature; low-latency serving of trained
models for use in real-time applications; seamless retraining and
redeployment of models as new data are available; and
model discovery, reuse, and citation.

The Data and Learning Hub for science (DLHub) is a learning system designed to
address inefficiencies in two important phases of the ML lifecycle, namely the
publication and serving of ML models plus associated data.
DLHub provides a nexus for ML practitioners and model
creators to publish and share models while capturing model provenance and
providing credit to contributors. It includes a flexible, user-managed 
model catalog built upon
a common metadata schema to enable discovery. 
DLHub also provides a low-latency and scalable model serving framework for executing
inference and other tasks on arbitrary data, and for linking multiple data processing and
model execution steps into end-to-end pipelines.
This framework can execute containerized DLHub models efficiently on a variety 
of parallel and distributed computing resources for both low latency and high throughput.

Using DLHub, users can securely share and publish models and transformation codes,
discover cutting-edge models published by the community, compose multi-step
inference pipelines with pre-/post-processing stages, and perform
inference at scale. Unlike other model repositories and serving systems, DLHub supports a
wide range of model types including TensorFlow~\cite{abadi2016tensorflow},
Keras~\cite{chollet2017deep}, and Scikit-learn~\cite{pedregosa2011scikit}; integrates with
Globus~\cite{chard2016globus} to provide seamless authentication and 
high performance data access for training
and inference; scales execution
environments dynamically to provide scalable and low-latency inference; 
and supports workflows that link
pre/post-processing steps and ML models. 

In the following,
we first motivate the need for science learning systems in \S\ref{sec:motivation} 
and survey model repository and serving frameworks in \S\ref{sec:survey}.
We then describe the DLHub architecture and implementation in \S\ref{sec:dlhub},
explaining how it extends prior research efforts, for example by
supporting both a wider variety of 
model types and the end-to-end workflows required for scientific uses of ML. 
In \S\ref{sec:evaluation}, we present experimental results 
which show that DLHub's scalable serving solution performs 
comparably with other systems, such as TensorFlow Serving~\cite{tfserving} and 
SageMaker~\cite{sagemaker}, and significantly better when memoization and batching
can be employed.
We show that without memoization, DLHub can serve requests to run models in less than 40ms and Python-based 
test functions in less than 20ms on a Kubernetes cluster.
Finally, we describe early uses
of DLHub in \S\ref{sec:usecases} and conclude in \S\ref{sec:conclusion}.

\section{Specialized Requirements of Science}
\label{sec:motivation}

Increasingly sophisticated learning systems are being developed, in particular by the major cloud
providers, to support commercial ML use cases.
However, scientific use of ML has specialized requirements, including the following.

\textbf{Publication, citation, and reuse:} The scholarly process is built upon a common
workflow of publication, peer review, and citation. Progress is dependent on 
being able to locate, verify, and build upon prior research, 
and careers are built upon publications and citation. As
scholarly objects, ML models should be subject to similar publication, review, 
and citation models on which the scholarly process is built. Lacking standard
methods for doing so, (a) many models associated with published
literature are not available~\cite{gundersen2017state}; and (b) researchers adopt
a range of ad hoc methods (from customized websites to GitHub) for sharing ML models~\cite{gossett2018aflowml, zhang2018ocpmdm, agrawal2018fatiguepredictor}.

\textbf{Reproducibility:}  Concerns about reproducibility are having a profound effect on 
research~\cite{baker16reproducibility}. While reproducibility initiatives have primarily focused on making data and 
experimental processes available to reproduce findings, there is a growing interest
in making computational methods available as well~\cite{morin2012shining,brinkman18wholetale, stodden16reproducibility}.  

Unlike sharing software products, there is little guidance for sharing ML models
and their artifacts (e.g., weights, hyper-parameters, and training/test sets).
Without publishing these artifacts, it is almost impossible to verify or build
upon published results.
Thus, there is a growing need to develop standard ML model packages
and metadata schema, and to provide rich model repositories and
serving platforms that can be used to reproduce published results.

\textbf{Research infrastructure:}
While industry and research share common requirements for scaling inference, the
execution landscape differs. Researchers often want to use multiple
(often heterogeneous) parallel and distributed computing resources to
develop, optimize, train, and execute models. Examples include: laboratory
computers, campus clusters, national cyberinfrastructure (e.g., XSEDE~\cite{towns2014xsede},
Open Science Grid~\cite{pordes2007open}),
supercomputers, and clouds. Thus, we require learning systems that can support
execution on different resources and enable migration between resources.

\textbf{Scalability:}
Large-scale parallel and distributed computing environments enable ML models to
be executed at unprecedented scale. Researchers require learning systems that
simplify training and inference on enormous scientific datasets and that can be
parallelized to exploit large computing resources.

\textbf{Low latency:} ML is increasingly being used in real-time scientific
pipelines, for example to process and respond to
events generated from sensor networks; classify and prioritize
transient events from digital sky surveys for exploration; and to perform error
detection on images obtained from X-ray light sources. There is a need in each
case for low latency, near real-time ML inference for anomaly/error detection
and for experiment steering purposes. As both the number of devices and
data generation rates continue to grow,
there is also a need to be able to execute
many inference tasks in parallel, whether on centralized or ``edge" computers.

\textbf{Research ecosystem:} Researchers rely upon a large and growing
ecosystem of research-specific software and services: for example,
Globus~\cite{chard2016globus} to access and manage their data; 
community and institution-specific data sources (e.g., the Materials Data
Facility~\cite{blaiszik2016materials} and Materials Project~\cite{jain2013commentary}) as input to 
their ML models;
and research authentication and authorization
models (e.g., using campus or ORCID identities).

\textbf{Workflows:} Scientific analyses often involve multiple steps, such as the staging of input 
data for pre-processing and normalization, extraction of pertinent features,
execution of one or
more ML models, application of uncertainty quantification methods,
post-processing of outputs, and recording of provenance. There is a growing need to
automate invocation of such sequences of steps into cohesive, shareable, and
reusable workflows. It is important, furthermore, that such processing workflows can be adapted easily to
meet the specific needs of a particular application.

\section{Learning Systems: A Brief Survey}
\label{sec:survey}

We define a learning system as ``a system that supports any phase of the
ML model lifecycle including the development, training, inference,
sharing, publication, verification, and reuse of a ML model.''
To elucidate the current landscape, we survey a range of existing systems,
focusing on those that provide model repository and serving capabilities.
Model repositories 
catalog collections of models, maintaining metadata for the purpose
of discovery, comparison, and use. Model serving platforms facilitate
online model execution.

A repository or serving platform may be provided as a \textit{hosted} service, in which case models are
deployed and made available to users via the Internet, or \textit{self-service},
requiring users to operate the system locally and manage the deployment of
models across their own infrastructure. 

While many learning systems provide
other capabilities (e.g., interactive computing interfaces via Jupyter
notebooks~\cite{jupyter}), we limit our survey to repository and serving
capabilities.

\subsection{Model Repositories}

Model repositories catalog and aggregate models, often by domain, storing
trained and untrained models with associated metadata to enable discovery and citation.
Metadata may be user-defined and/or standardized by using common publication
schemas (e.g., author, creation date, description, etc.) and ML-specific
schemas.  ML-specific metadata include
model-specific metadata (e.g., algorithm, software version, network architecture),
development provenance (e.g., versions, contributors),
training metadata (e.g., datasets and parametrization used for
training), and performance metadata (e.g., accuracy when
applied to benchmark datasets).
Model repositories may provide the ability to associate a persistent identifier
(e.g., DOI) and citation information such that creators may receive credit for
their efforts.

Table~\ref{table:repo_summary} summarizes four representative model repositories plus DLHub along
the following dimensions; we describe each repository in more detail below.

\begin{itemize}
 \item \textbf{Publication and curation:} Can models be contributed by users and is any curation process applied.
 \item \textbf{Domain:} Whether the repository is designed for a single domain (e.g., bioinformatics) or for
many domains.
 \item \textbf{Model types:} What types of ML models can be registered in the repository (e.g., any 
model type, TensorFlow)
 \item \textbf{Data integration:} Whether data (e.g., training/test datasets) and configuration
(e.g., hyperparameters) can be included with the published model.
 \item \textbf{Model metadata:} Whether the repository supports publication, model-specific, model 
building, and/or invocation metadata. 
 \item \textbf{Search capabilities:} What search mechanisms are provided to allow users to find and 
compare models.
 \item \textbf{Model versioning:} Whether the repository facilitates versioning and updates to published models.
 \item \textbf{Export}: Does the repository allow models to be exported and if so in what format.
\end{itemize}

\definecolor{Gray}{gray}{0.9}
\newcolumntype{a}{>{\columncolor{Gray}}c}

\begin{center}
	\begin{table*}[t]
    \centering
    \vspace{-0mm}
	\caption{Model repositories compared and contrasted. BYO = bring your own.}
\vspace{-0mm}
\begin{tabular}{ | l | c| c | c|c|a|}
\hline
 & \textbf{ModelHub} & \textbf{Caffe Zoo} & \textbf{ModelHub.ai} & \textbf{Kipoi} & \textbf{DLHub}  \\ \hline
\textbf{Publication method} & BYO & BYO & Curated & Curated & BYO \\ \hline
\textbf{Domain(s) supported} & General & General & Medical & Genomics & General \\ \hline
\textbf{Datasets included} & Yes & Yes & No & No & Yes \\ \hline
\textbf{Metadata type} & Ad hoc & Ad hoc & Ad hoc & Structured & Structured \\ \hline
\textbf{Search capabilitiees} & SQL & None & Web GUI & Web GUI & Elasticsearch \\ \hline
\textbf{Identifiers supported} & No & BYO & No & BYO & BYO \\ \hline
\textbf{Versioning supported} & Yes & No & No & Yes & Yes \\ \hline
\textbf{Export method} & Git & Git & Git/Docker & Git/Docker & Docker \\ \hline
\end{tabular}
	\label{table:repo_summary}
    \vspace{-0mm}\vspace{-0in}
	\end{table*}
\end{center}

\vspace{-4ex}

\subsubsection{ModelHub}
~\cite{miao2017towards} is a deep learning model lifecycle management
system focused on managing the data artifacts generated during the deep learning
lifecycle, such as parameters and logs, and understanding the behavior of the
generated models. Using a Git-like command line interface, users initialize
repositories to capture model information and record the files created during
the creation process. Users then exchange a custom-built model versioning
repository, called DLV, through the hosted service to enable publication and
discovery. ModelHub is underpinned by Git, inheriting versioning capabilities,
support for arbitrary datasets, scripts, and features, and accommodates models
regardless of domain. A custom SQL-like query language, called DQL, allows
ModelHub users to search across repositories by
characteristics such as authors, network architecture, and hyper-parameters.

\subsubsection{Caffe Model Zoo}
~\cite{CaffeModelZoo} is a community-driven effort to
publish and share Caffe~\cite{jia2014caffe} models. Users contribute
models via Dropbox or Github Gists. The Model Zoo provides a standard format for
packaging, describing, and sharing Caffe models. It also provides tools
to enable users to upload models and download trained binaries. The Model Zoo
operates a community-edited Wiki page to describe each of the published models,
aggregating information regarding manuscripts, citation, and usage
documentation in an unstructured format.
The project encourages open sharing of models, trained weights,
datasets, and code through Github. The Model Zoo
provides guidelines on how
to contribute models and what metadata should be included in the accompanying
\textit{readme.md} file without enforcing a specific schema. Users typically include citation 
information (e.g., BibTeX
references to papers), links to the project page, a Github address for the
model's code, and in some cases, a link to \texttt{haystack.ai} where the model can be
tested.

\subsubsection{ModelHub.ai}
~\cite{ModelHubai} is a service to crowdsource and aggregate deep learning models
related to medical applications. ModelHub.ai has a Web interface that
lets users review published models, experiment with example inputs, and
even test them online using custom inputs. The service provides detailed
documentation and libraries to package models into a supported Docker format.
Once packaged, users can add the model and any associated metadata
to the ModelHub GitHub repository and submit a pull-request.
The contributed model is curated and added to the catalog.
The ModelHub.ai project provides both a Flask and Python
API to interact with Dockerized models, which can be retrieved by either downloading the Docker image or cloning the GitHub repository.

\subsubsection{Kipoi}~\cite{avsec2018kipoi} 
is a repository of trained models for genomics that includes
more than 2000 models of 21 different types.
It provides a command line interface (CLI) for publishing and accessing models.
On publication, the CLI queries the user for descriptive metadata and generates a configuration file containing
the metadata needed to discover and run the model.
Users can then publish their models by submitting a
pull-request to the Kipoi GitHub repository. 
Models can be listed and retrieved through the API and then invoked locally.

\subsection{Model Serving}

ML model serving platforms provide on-demand model inference.
Existing model serving platforms vary in both their goals 
and capabilities: for example, some focus on serving a specific type of model with extremely low latency, 
while others prioritize ease of use and simple inference interfaces. 
We have identified the following important dimensions to capture
the differences between model serving platforms.
Table~\ref{table:serving_summary} summarizes popular model serving platforms plus DLHub
along these dimensions.

\begin{itemize}
 \item \textbf{Service model:} Whether the platform is offered as a hosted service or
requires self-service deployment.
 \item \textbf{Model types:} What languages and types are supported (e.g., TensorFlow, Scikit-learn,
R, Python, etc.)
 \item \textbf{Input types:} The range of input types supported by the system. (e.g., structured,
files, or primitive types)
 \item \textbf{Training capabilities:} Whether the system supports training.
 \item \textbf{Transformations:} Whether pre-/post-processing steps can be
deployed.
 \item \textbf{Workflows:} Support for workflows between models and transformation codes.
 \item \textbf{Invocation interface:} Methods of interaction with the models.
 \item \textbf{Execution environment:} Where models are deployed (e.g., cloud, Kubernetes, Docker).
\end{itemize}

\begin{center}
	\begin{table*}[t]
    \centering
    \vspace{-0mm}
	\caption{Serving systems compared and contrasted. K8s = Kubernetes.}
\vspace{-0mm}
\begin{tabular}{ | l | c| c | c|c|a|}
\hline
 & \textbf{PennAI} & \textbf{TF Serving} & \textbf{Clipper} & \textbf{SageMaker} & \textbf{DLHub} \\ \hline
\textbf{Service model} & Hosted & Self-service & Self-service & Hosted & Hosted \\ \hline
\textbf{Model types} & Limited & TF Servables & General & General & General \\ \hline
\textbf{Input types supported} & Unknown & Primitives, Files & Primitives & Structured, Files &
Structured, Files \\ \hline
\textbf{Training supported} & Yes & No & No & Yes & No \\ \hline
\textbf{Transformations} & No & Yes & No & No & Yes \\ \hline
\textbf{Workflows} & No & No & No & No & Yes \\ \hline
\textbf{Invocation interface} & Web GUI & gRPC, REST & gRPC, REST & gRPC, REST & API, REST \\ \hline
\textbf{Execution environment} & Cloud & Docker, K8s, Cloud & Docker, K8s & Cloud, Docker & K8s,
Docker, Singularity, Cloud \\ \hline
\end{tabular}
	\label{table:serving_summary}
    \vspace{-0mm}\vspace{-0in}
	\end{table*}
\end{center}

\vspace{-4ex}

\subsubsection{PennAI}~\cite{olson2018system} provides model serving capabilities for 
biomedical and health data. The platform allows users to apply six ML algorithms, including regressions, 
decision trees, SVMs, and random forests to their datasets, and to perform 
supervised classifications. The PennAI website provides a user-friendly interface for selecting, 
training, and applying algorithms to data. The platform also exposes a controller for 
job launching and result tracking, result visualization tools, and a graph database to store 
results. PennAI does not support user-provided models, but does provide an intuitive 
mechanism to train classification tools and simplify the integration of ML into scientific 
processes.

\subsubsection{TensorFlow Serving}

~\cite{olston2017tensorflow} is the most well-known model
serving solution and is used extensively in production environments. TensorFlow Serving
provides high performance serving via gRPC and REST APIs and is capable of
simultaneously serving many models, with many versions, at scale. TensorFlow Serving
provides the lowest latency serving of any of the surveyed platforms.
It serves trained TensorFlow models using the standard
\texttt{tensorflow\_model\_server}, which is built in C++. Although TensorFlow Serving does
support a range of model types---those that can be exported into TensorFlow servables---it is limited in terms of its support for custom transformation codes and does not
support the creation of pipelines between servables. TensorFlow Serving is also
self-service, requiring users to deploy and operate TensorFlow Serving on local (or
cloud) infrastructure in order to deposit models and perform inferences. TensorFlow also provides model 
repository capabilities through a library of reusable ML modules, called TensorFlow Hub.

\subsubsection{SageMaker}

~\cite{sagemaker} is a ML serving platform provided by
Amazon Web Services that supports both the training of models and the
deployment of trained models as Docker containers for
serving. It helps users to handle large data efficiently by providing ML algorithms 
that are optimized for distributed environments.
SageMaker APIs allow users
to deploy a variety of ML models and integrate their own
algorithms. 
In addition, trained models can be exported as Docker containers for local 
deployment.

\subsubsection{Clipper}

~\cite{crankshaw2017clipper} is a prediction serving system that focuses
on low latency serving.
It deploys models as Docker containers, 
which eases management complexity
and allows each model to have its own dependencies wrapped in a self-contained
environment. Clipper includes several optimizations to improve
serving performance including data batching and memoization. Clipper also provides a
model selection framework to improve prediction accuracy. However, because Clipper
needs to dockerize the models on the manager node, it requires
privileged access, which is not available on all execution environments (e.g.,
high performance computing clusters).

\subsubsection{Kubeflow}
~\cite{Kubeflow} is a collection of open source ML services that can be
deployed to provide a fully-functional ML environment on a Kubernetes cluster.
The system simplifies the deployment of various ML tools and services, including
those to support model training, hyper-parameter tuning, and model serving. 
Kubeflow also integrates Jupyter Notebooks to provide a user-friendly
interface to many of these services. Model serving in Kubeflow is achieved
with TensorFlow Serving, therefore we have not included it in our summary table. However, using 
TensorFlow Serving provides Kubeflow with a low-latency model serving
solution while adding additional capabilities (e.g., training) to the platform.

\section{DLHub Architecture and Implementation}
\label{sec:dlhub}

\begin{figure}[h]
 \centering
 \includegraphics[width=\columnwidth,trim=0.1in 4.1in 6.5in 0.2in,clip]{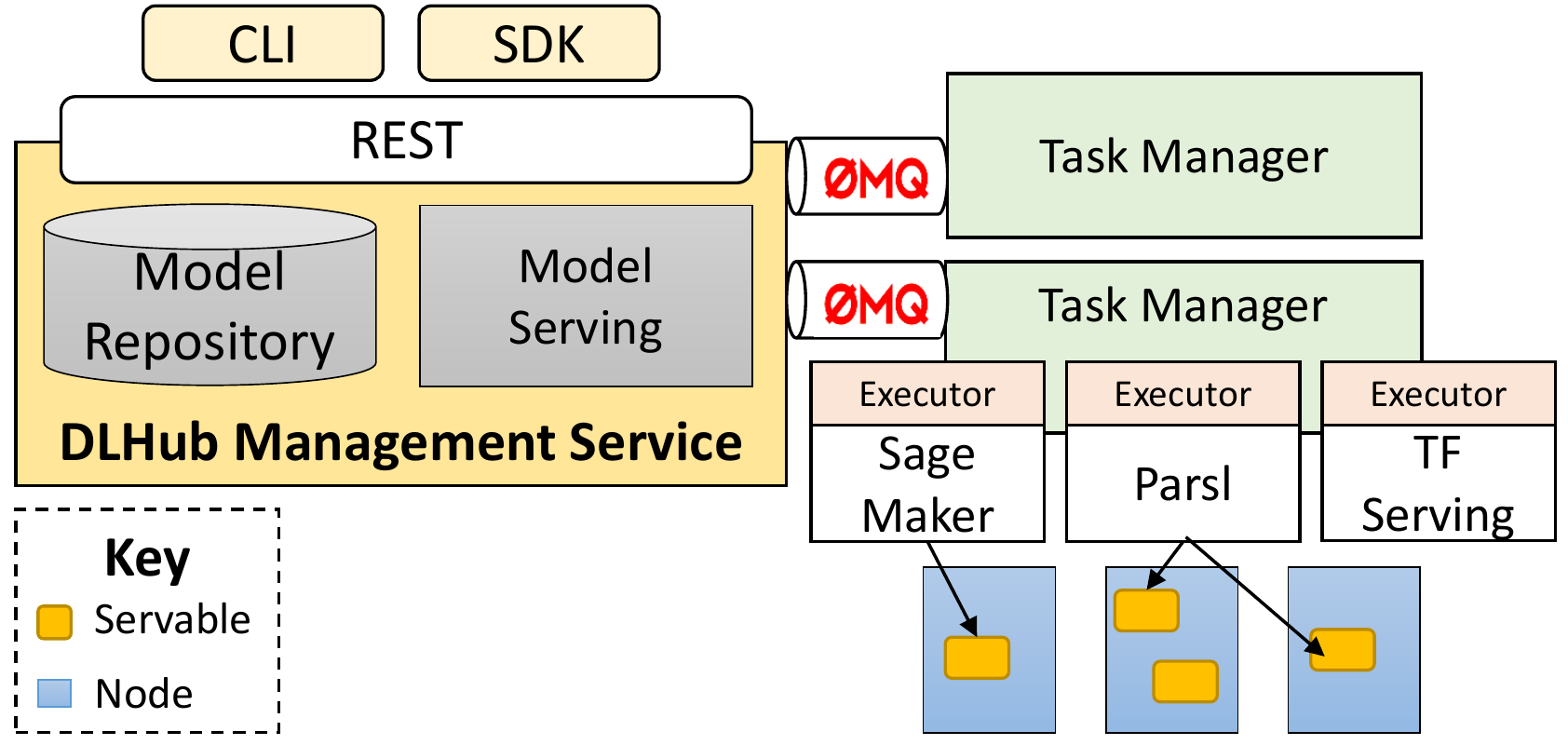}
 \caption{DLHub architecture.
 User requests, submitted via REST, SDK, or CLI (upper left) can result in
 model publication in the Repository or the dispatch of serving
 requests to servables deployed on any computing resources with a Task Manager
 interface and appropriate executor(s) (lower right).}
 \label{fig:arch}
\end{figure}

DLHub is a learning system that provides model publishing and serving capabilities 
for scientific ML. 
DLHub's model repository supports user-driven publication, 
citation, discovery, and reuse of ML models from a wide range of domains. 
It offers rich search capabilities to enable discovery of, and access to, published models.
DLHub automatically converts each published model
into a ``servable''---an executable DLHub 
container that implements a standard execution interface and comprises a complete model package
that includes the trained model, model components (e.g., training 
weights, hyperparameters), and any dependencies (e.g., system or Python packages). 
DLHub can then ``serve'' the model by deploying and invoking one or more instances of the servable 
on execution site(s).
DLHub provides high throughput and low-latency model serving by dispatching
tasks in parallel to the remote execution site(s). DLHub implements a flexible 
\emph{executor} model in which several serving infrastructures, 
including TensorFlow Serving, SageMaker, and our general-purpose 
Parsl-based execution platform, can be used to execute tasks.
The DLHub architecture, shown in \figurename{~\ref{fig:arch}}, comprises three
core components: 
the \textit{Management Service}, one or more \textit{Task Managers}, and 
one or more \textit{Executors}.

\subsection{Management Service}

DLHub's Management Service is the user-facing interface to DLHub. 
It enables users to publish models, query available models, execute 
tasks (e.g., inference), construct pipelines, and monitor the status of tasks. 
The Management Service includes advanced functionality to build models, optimize 
task performance, 
route workloads to suitable executors, batch tasks, and cache results.
 
\textbf{Model Repository:}
A primary function of the Management Service is to implement model repository functionality
to support the publication and discovery of models.
DLHub defines a model publication schema
that is used to describe all published models. The schema includes 
standard publication metadata (e.g., creator, date, name, description)
as well as ML-specific metadata such as
model type (e.g., Keras, TensorFlow) and input and output data types. 
These metadata are registered in a Globus Search~\cite{ananthakrishnan18publication} 
index that can be
queried by users to discover appropriate models. 

\textbf{Model discovery:}
DLHub's search interface supports fine-grained, access-controlled queries
over model metadata. When a new model is published
in DLHub, its metadata are indexed in Globus Search. It can the be queried
using free text queries, partial matching, range queries, faceted search, and more.

\textbf{Servables:}
In order to provide a common model execution interface irrespective of model
type, DLHub converts all published models into executable servables. 
Thus, DLHub requires that users upload
all model components as well as descriptive metadata 
for building the servable. As model components can be large, model 
components can be uploaded to an AWS S3 bucket or a Globus endpoint. 
Once a model is published, the Management Service downloads the components 
and builds the servable in a DLHub-compatible format. It
combines DLHub-specific dependencies with user-supplied model dependencies
into a Dockerfile. It then uses the Dockerfile to create a Docker container
with the uploaded model components and all required dependencies. Finally, it
uploads the container to the DLHub 
model repository, and
registers the container location in the Globus Search index alongside
the user-supplied descriptive metadata.

\textbf{Model serving:}
The Management Service also coordinates the execution of tasks on
remote resources. It uses a ZeroMQ~\cite{hintjens2013zeromq} queue
to send tasks to registered Task Managers for execution. 
The queue provides a reliable
messaging model that ensures tasks are received and executed. 
When a user invokes a model (or executes another task), 
with some input data, the Management Service packages up the request
and posts it to a ZeroMQ queue. Registered Task Managers can retrieve 
waiting tasks from the queue, unpackage the request, execute the task, 
and return the results via the same queue. DLHub supports both synchronous
and asynchronous task execution. In asynchronous mode, the Management
Service returns a unique task UUID that can be used subsequently to monitor
the status of the task and retrieve its result. 

\subsection{Task Managers}
Any compute resource on which DLHub is to execute tasks must be 
preconfigured with DLHub Task Manager software. The Task Manager
is responsible for monitoring the DLHub task queue(s) and then executing
waiting tasks. The Task Manager is responsible for deploying servables
using one of the supported executors and then routing tasks to appropriate
servables. 
When a Task Manager is first deployed it registers itself with the Management Service and 
specifies which executors and DLHub servables it can launch.  
The Task Manager can be deployed in Docker environments, Kubernetes clusters, and HPC 
resources via Singularity. 

\subsection{Executors}
DLHub aims to provide efficient model execution for a wide range
of model types. To achieve this goal it implements an arbitrary
executor model that currently supports three serving systems: 
TensorFlow Serving, SageMaker, and a general-purpose
Parsl executor.
The first two executors work only with containers deployed to Kubernetes, while
the Parsl executor can support Kubernetes and many other common HPC schedulers and clouds.

When invoking a given servable the Task Manager determines to which
executor to send the task. Inference tasks are sent to the appropriate
serving executor, other types of tasks (e.g., pre/post processing) are
sent to the general Parsl executor. Each executor is responsible for running
the task, translating the results into a common DLHub executor-independent input
format, and returning the results back to the Management Service via
the Task Manager. 
While servables might be invoked through different executors, all
requests are managed by the Management Service 
and benefit from the same
batching, memoization, and workflow capabilities,

\textbf{TensorFlow Serving executor:} To invoke TensorFlow
models, the Task Manager first deploys one or more TensorFlow Serving 
containers on Kubernetes. 
It then routes subsequent TensorFlow servable 
invocations to these containers. 
The executor uses Google's low-latency gRPC protocol
for high performance model inference. 

\textbf{SageMaker executor:}  The SageMaker executor deploys
one or more SageMaker containers on Kubernetes. The SageMaker
container includes a Python Flask
application that exposes an HTTP-based model inference interface. The Task
Manager composes HTTP requests to the SageMaker interface
to perform inference. 

\textbf{Parsl executor:}
The Parsl executor uses the Parsl~\cite{parsl} parallel scripting library's
execution engine to execute tasks on arbitrary resources.
This execution engine supports the execution of Python functions,  
arbitrary executables, and containerized applications on different platforms, via a modular
execution interface that allows a client (in our case, the Task Manager) to specify the desired execution location
and associated parameters, such as degree of parallelism. 

Implementations of the Parsl execution engine on cluster, cloud, and supercomputer platforms,
among others, use platform-specific mechanisms to dispatch tasks and their data, initiate task execution,
monitor progress, and report on results.
On a Kubernetes cluster, for example,
the engine creates 
a Kubernetes Deployment consisting of \emph{n} pods for each servable that is to be executed,
a number configurable in the Management Service. Parsl then deploys IPythonParallel (IPP) engines in 
each servable container and connects back to the Task Manager to retrieve servable execution requests.  
Parsl dispatches requests to the appropriate containers using IPP, load balancing them
automatically across the available pods.

\subsection{Security}
DLHub implements a comprehensive security model that ensures that all operations are performed
by authenticated
and authorized users. To do so, we rely on Globus Auth---a flexible authentication
and access management service that is designed to broker authentication and authorization
decisions between users, identity providers, resource serves, and clients. 

The DLHub Management Service is registered as a Globus Auth resource server~\cite{GlobusAuth}
with associated scope for programmatic invocation. Users can authenticate with DLHub
using one of hundreds of supported identity providers (e.g., campus, ORCID, 
Google). When authenticating, the Management Service first validates the 
user's identity, and then retrieves short-term access tokens
that allow it to obtain linked identities, profile information about the user, 
and to access/download data on their behalf. 
These capabilities allow DLHub to precomplete
publication metadata using profile information and also to transfer model components and inputs
from Globus endpoints seamlessly.

\subsection{DLHub interfaces}

DLHub offers a REST API, Command Line Interface (CLI), and a Python Software Development 
Kit (SDK) for publishing, managing, and invoking models. 
We also provide a user toolbox to assist with the creation of 
metadata that adhere to DLHub model schemas. 

The DLHub \textbf{toolbox} supports programmatic construction of JSON documents
that specify publication and model-specific metadata that complies with DLHub-required schemas.
The resulting documents can then be uploaded to DLHub to publish the model.
The toolbox also provides functionality to execute DLHub models locally. 
This functionality is useful for model development and testing. 

The DLHub \textbf{CLI} provides an intuitive user interface to interact with DLHub. 
It provides a Git-like interface with commands
for initializing a DLHub servable in a local directory, publishing the servable to 
DLHub, creating metadata using the toolbox, and invoking the published servable
with input data. Supported commands include:

\begin{itemize}
 \item \textbf{init}: Initialize a servable in the current working directory on the user's 
computer, 
creating a  \texttt{.dlhub} directory and a metadata file for the servable.
 \item \textbf{update}: Modify a servable's published metadata.
 \item \textbf{publish}: Push a locally created servable (model and metadata) to DLHub. 
 \item \textbf{run}: Invoke a published servable with appropriate input arguments. 
 \item \textbf{ls}: List the servables that are being tracked on the user's computer.
\end{itemize}

The DLHub \textbf{Python SDK} supports programmatic interactions with DLHub. 
The SDK wraps DLHub's REST API, providing access to all model repository and serving 
functionality. 
\section{Evaluation}
\label{sec:evaluation}

To evaluate DLHub we conducted experiments to explore its serving
performance, the impacts of memoization and data batching,
and Task Manager scalability.
We also compared its serving infrastructure
against those of TensorFlow Serving, Clipper, and SageMaker.

\subsection{Experimental Setup}

We use Argonne National Laboratory's PetrelKube,
a 14-node Kubernetes cluster, as a compute resource for these experiments. 
Each node has two E5-2670 CPUs, 128GB RAM,
two 300GB hard drives in RAID 1,
two 800GB Intel P3700 NVMe
SSDs, and 40GbE network
interconnection.

In our experiments, we deployed a single Task Manager on a co-located cluster, Cooley, within Argonne's
Leadership Computing Facility.
The average Internet Protocol round-trip-time between the Task Manager and
PetrelKube, where the servables are deployed, is 0.17ms. The
Management Service was deployed on an Amazon EC2 instance and has an average round-trip-time to the
Task Manager of 20.7ms.
These overheads are consistent across our tests and are present regardless of
executor or serving infrastructure used.

When reporting the end-to-end performance of servable invocations we use the following metrics.
\begin{itemize}
\item{\bf Inference time} is captured at the \emph{servable} and measures the time taken by the
servable to run the component.
\item{\bf Invocation time} is captured at the \emph{Task Manager} and measures elapsed time
from when a request is made to the executor to when the result is received from the servable.
\item{\bf Request time} is captured at the Management Service and measures the time from receipt of the task request to receipt of its result from the Task Manager.
\item{\bf Makespan} is the completion time of serving all the submitted requests.
\end{itemize}

We use six servables to compare the different serving infrastructures. 
The first is a baseline 
``\textbf{noop}'' task that returns ``hello world'' when invoked. 
The second is Google's 22-layer 
Inception-v3~\cite{szegedy2016rethinking} model
(``\textbf{Inception}"). Inception is trained on a large academic dataset for image
recognition and classifies images into 1000 categories. Inception takes
an image as input and outputs the five most likely categories.

The third is a multi-layer convolutional neural network trained on
CIFAR-10~\cite{krizhevsky2009learning} (``\textbf{CIFAR-10}"). 
This common benchmark
problem for image recognition takes a 32$\times$32 pixel RGB image as input and 
classifies it in 10 categories.

The final three servables are part of a workflow
used to predict the stability of a material given its
elemental composition (e.g., NaCl). The model is split into three servables: parsing a string with pymatgen~\cite{ong2013python} to extract
the elemental composition (``\textbf{matminer\_util}''), computing
features from the element fractions by using Matminer~\cite{ward2018matminer}
(``\textbf{matminer\_featurize}''), and executing a scikit-learn random forest model to predict
stability (``\textbf{matminer\_model}''). The model was trained with the features of Ward et
al.~\cite{ward2016magpie} and data from the Open Quantum Materials Database~\cite{Kirklin2015}.

\subsection{Experiments}

To remove bias we disable DLHub memoization mechanisms
and restrict data batch size to one in the following experiments, except where otherwise noted. 
Although both the Management Service and Task Manager
are
multi-threaded, we conduct all experiments sequentially, waiting for a response before submitting
the next request.

\subsubsection{Servable Performance.}~\label{sec:local}
To study the performance of the DLHub components (i.e., servable, Management Service, and Task
Manager) we
measure request, invocation, and inference times when requests are issued via the Management
Service.
We evaluate six servables on PetrelKube, submitting 100 requests with fixed input data to each.

\begin{figure}[h]
 \centering
 \includegraphics[width=\columnwidth,trim=0.1in 0in 0.6in 0.4in,clip]{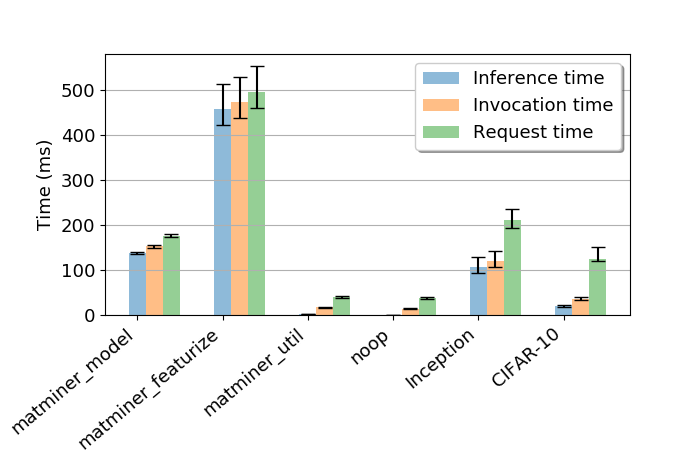}
 \caption{Request, invocation, and inference times for six servables. Bars give median values and 
the error bars the 5th and 95th percentile values.
}
 \label{fig:local}
\end{figure}

\figurename{~\ref{fig:local}} shows the inference,
invocation, and request times for each of the six servables. 
The difference between inference and invocation times reflects costs associated with the Task Manager while
the difference between invocation and request times reflects costs associated with the Management Service.
In most cases, costs are around 10--20ms; as these include round-trip-time
communication times, it seems that DLHub imposes only modest additional overhead. 
The higher overheads associated with Inception and CIFAR-10 are
due to their need to transfer substantial input data.

\subsubsection{Memoization.}
\begin{figure}[h]
 \centering
 \includegraphics[width=\columnwidth,trim=0.2in 0in 0.6in 0.2in,clip]{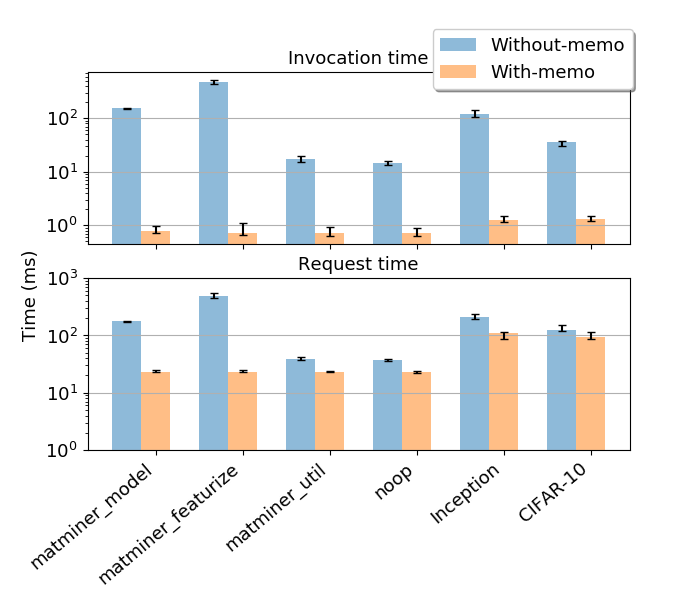}
 \caption{Performance impact of memoization. Bars and error bars show median and 5th/95th
percentiles, as in \figurename~\ref{fig:local}.}
 \label{fig:caching}
\end{figure}

~DLHub's Parsl executor implements memoization~\cite{michie1968memo},
caching the inputs and outputs for each request and returning the recorded output
for a new request if its inputs are in the cache.
To investigate the effect of memoization on
serving performance we submitted requests with the same input as
in \S\ref{sec:local}, with memoization variously enabled and disabled.
\figurename{~\ref{fig:caching}} shows invocation and request times.
Inference time is not shown in the figure as memoization removes the need to execute the inference task.
We observe that memoization reduces invocation time by 95.3--99.8\% and
request time by 24.3--95.4\%.

\subsubsection{Batching.}
\begin{figure}[h]
 \centering
 \includegraphics[width=\columnwidth,trim=0.15in 0.05in 0.6in 0.25in,clip]{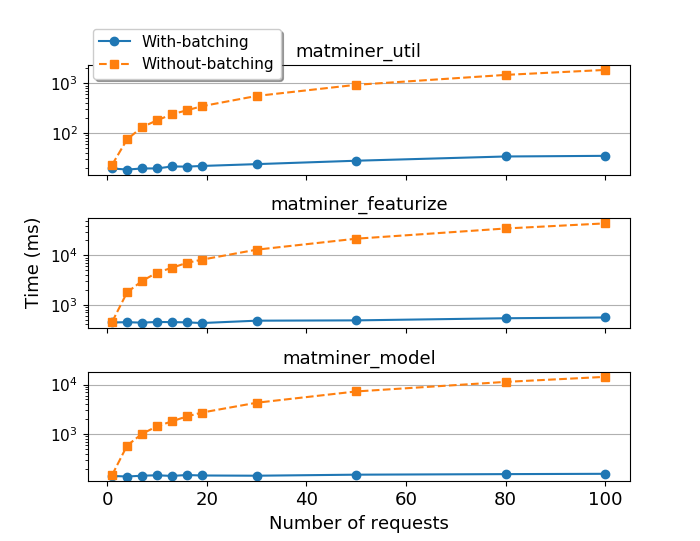}
 \caption{Servable invocation time, with and without batching.
}
 \label{fig:batching}
\end{figure}

\begin{figure}[h]
 \centering
 \includegraphics[width=\columnwidth,trim=0in 0.25in 0.45in 0.35in,clip]{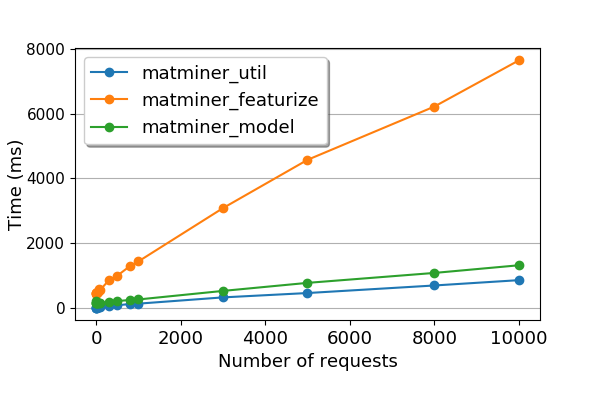}
 \caption{Invocation time vs.\ number of requests, with batching.}
 \label{fig:batching-model}
\end{figure}

~DLHub support for batch queries is designed to improve overall throughput by amortizing system overheads over many requests.
We studied how performance varies with batch size.
First, we measured the time required to process numbers of requests in the range [1, 100]
for  three example servables, with and without batching.
Due to space limitations, we show only invocation time in \figurename{~\ref{fig:batching}}.
We see that batching significantly reduces overall invocation time.
Next, we measured invocation time for the same three servables with batching,
as the number of requests scales far higher, to 10,000.
We see in \figurename{~\ref{fig:batching-model}} a roughly linear relationship between
invocation time and number of requests.
In future work, we intend to use such servable
profiles to design adaptive batching algorithms that intelligently distribute serving requests
to reduce latency.

\subsubsection{Scalability.}
\begin{figure}[h]
 \centering
 \includegraphics[width=\columnwidth,trim=0.15in 0.1in 0.5in 0.4in,clip]{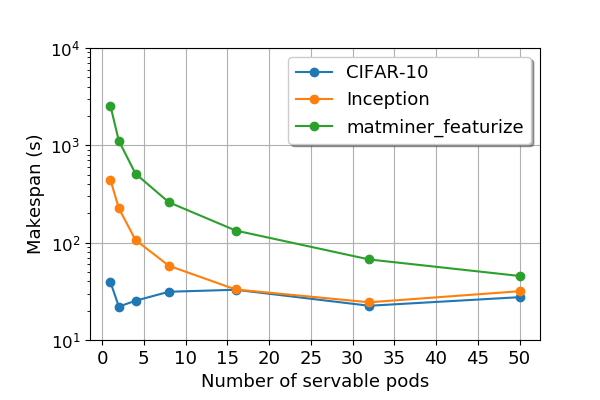}
 \caption{The time required for three different models to
process 5,000 inferences for varying numbers of replicas.}
 \label{fig:scaling}
\end{figure}

~We performed throughput tests with the Parsl executor, with
memoization disabled and a batch size of one. \figurename{~\ref{fig:scaling}} shows observed
Task Manager throughput for the Inception, CIFAR-10, and Matminer\_featurize models as the
number of deployed model replicas is increased.
We find that behavior varies with model. For example, when serving Inception requests, 
throughput increases rapidly up to $\sim$15 replicas, after which 
subsequent replicas have diminishing effect and executor throughput eventually saturates.
As we expect, servables that execute for shorter periods benefit less
from additional replicas, presumably because
task dispatch activities eventually come to dominate execution time.

\subsubsection{Serving Comparison.}~
We used CIFAR-10 and Inception to compare the serving performance of TensorFlow Serving, SageMaker, Clipper, and DLHub.
For TensorFlow Serving, we
export the trained models and use
the standard \texttt{tensorflow\_model\_server}. For SageMaker, we use the SageMaker
service to create the models, which we then deploy in DLHub.
For Clipper, we use its Kubernetes container manager to deploy it on PetrelKube and
register the CIFAR-10 and Inception models with it.
For DLHub, we serve the two models with the Parsl executor.
We evaluated serving performance by submitting requests
to the Management Service and having the
Task Manager route
requests to each platform. We report the average time from 100 requests for each model and platform.

TensorFlow Serving provides two model serving APIs: REST and gRPC.
SageMaker also supports serving TensorFlow models through TensorFlow Serving
or its native Flask framework.
In our experiments, we explore all possible APIs and frameworks, i.e.,
TFServing-REST, TFServing-gRPC, SageMaker-TFServing-REST, SageMaker-TFServing-gRPC, SageMaker-Flask.
In addition, as DLHub and Clipper both support memoization, we evaluate those two systems
with and without memoization.

\begin{figure}[h]
 \centering
 \includegraphics[width=\columnwidth,trim = 0.05in 0.1in 0.35in 0.5in,clip]{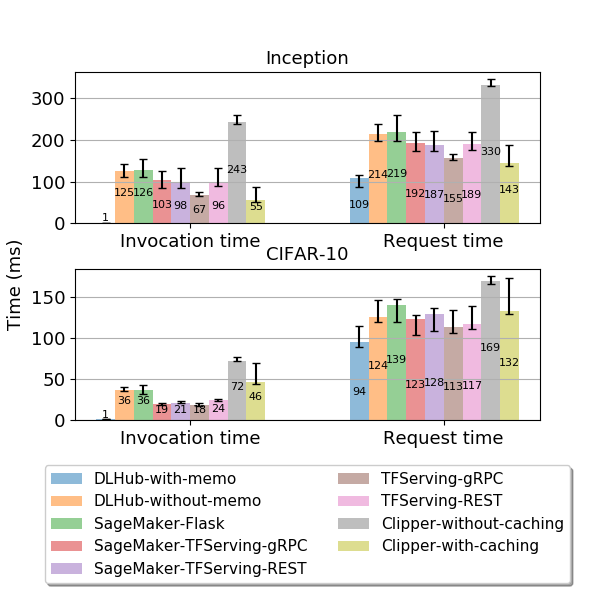}
 \caption{Performance of different serving systems on the Inception and CIFAR-10 problems.}
 \label{fig:comparison-serving}
\end{figure}

Figure~\ref{fig:comparison-serving} shows the invocation times and request times of CIFAR-10 and
Inception using each serving system.
We see that the servables invoked through the TensorFlow Serving framework (i.e., TFServing-gRPC,
TFServing-REST and SageMaker-TFServing) outperform those using other serving
systems (SageMaker-Flask and DLHub) in terms of both invocation time and request time.
This is because the core tensorflow\_model\_server, implemented in C++,
outperforms Python-based systems. DLHub's performance is comparable to the other Python-based
serving infrastructures.
In addition, gRPC leads to slightly better performance than REST due to the overhead of the
HTTP protocol. With memoization enabled, DLHub provides extremely low invocation times
(1ms)---much lower than other systems. 
This is because Parsl maintains a cache at the
Task Manager, greatly reducing serving latency. 
Clipper, in contrast,
maintains a cache at the query frontend that is deployed as a pod on the Kubernetes cluster
(PetrelKube). Hence, cached responses still require the request to be transmitted to the query
frontend, leading to additional overhead.

\section{Use Cases}
\label{sec:usecases}

To illustrate the value of DLHub we briefly outline 
four use cases that exemplify early adoption of the system. 

\subsection{Publication of Cancer Research Models}
The 
Cancer Distributed Learning Environment (CANDLE) 
project~\cite{wozniak2017candle}
leverages leadership scale computing resources to address problems relevant to cancer research at different 
biological scales, specifically problems at the molecular, cellular, and population scales. 
CANDLE uses DLHub to securely share 
and serve a set of deep learning models and benchmarks using cellular level data 
to predict drug response based on molecular features of tumor cells and drug descriptors.
As the models are still in development, they require substantial 
testing and verification by a subset of selected users prior to their general release. DLHub 
supports this use case by supporting model sharing and discovery with
fine grain access control. Thus, 
only permitted users can discover and invoke the 
models through the platform. Once models are determined suitable for general release, the access 
control on the model can be updated within DLHub to make them publicly available.

\subsection{Enriching Materials Datasets}
The Materials Data Facility~\cite{blaiszik2016materials} (MDF) is a set of data services developed
to enable data publication and data discovery in the materials science
community. MDF allows researchers to distribute their data, which may be large
or heterogeneous, and rapidly find, retrieve, and combine the contents of
datasets indexed from across the community.
MDF 
leverages several models published in DLHub to add value to
datasets as they are ingested. When a new dataset is registered with MDF, automated workflows~\cite{ananthakrishnan18publication} are applied to trigger the invocation of relevant
models to analyze the dataset and generate additional metadata. The selection of
appropriate models is possible due to the descriptive schemas used in both MDF
and DLHub. MDF extracts and associates fine-grained type information with each
dataset which are closely aligned with the applicable input types described for
each DLHub model.

\subsection{Processing Tomographic Neuroanatomy Data}
DLHub is used by a research group using X-ray microtomgraphy
at the Advanced Photon Source to 
rapidly characterize the
neuroanatomical structure of large (cm$^3$) unsectioned brain volumes in order
to study brain aging and disease~\cite{chard2018high}. During the image reconstruction
process, it is important to identify the center of the imaged sample from the 
hundreds of slices with varying quality.
A DLHub model is used to aid in the identification of the highest quality slice to
be used for tomographic reconstruction. Once reconstructed, the resulting images are
further processed with segmentation models to characterize cells. These results
can then be used to visualize and analyze the composition of the brain tissue.
DLHub hosts both the center-finding and segmentation models, enabling near
real-time automated application of the center finding models during the
reconstruction process as well as facilitating batch-style segmentation
post-processing of images.

\subsection{Predicting Formation Enthalpy}
DLHub makes it easy to link models plus pre- and post-processing transformations
in pipelines to simplify the user experience. For example, a pipeline for
predicting formation enthalpy from a material composition (e.g., SiO$_2$) can be
organized into three steps: 1) conversion of material composition text into a
pymatgen~\cite{ong2013python} object; 2) creation of a set of features, via
matminer~\cite{ward2018matminer}, using the pymatgen object as input; and 3)
prediction of formation enthalpy using the matminer features as input. Once the
pipeline is defined, the end user sees a simplified interface that allows them to input a
material composition and receive a formation enthalpy. This and other more complex
pipelines are defined as a series of
modularized DLHub servables 
within the
execution environment. Defining these steps as a pipeline
means data are automatically passed between each servable in the
pipeline, meaning the entire execution is performed server-side, drastically
lowering both the latency and user burden to analyze inputs. 

\section{Conclusion}
\label{sec:conclusion}

The increasing reliance on ML in science necessitates the development of new
learning systems to support the publication and serving of models. 
To address these requirements we have developed DLHub, a
unique model repository and serving system
that provides standardized, self-service model
publication, discovery, and citation; scalable execution of 
inference tasks on arbitrary execution resources; integration with
large-scale research storage systems; and support for flexible 
workflows. We showed that DLHub performs comparably with other serving platforms,
and significantly better when its  
memoization and batching capabilities can be used, and can provide low latency inference. 

In ongoing research, we are working to expand the set of models included in DLHub, 
simplify use by domain scientists, and study how DLHub capabilities are used in practice. 
We are investigating the utility of integrating multiple servables
into single containers and optimization techniques for automated tuning of
servable execution. 
We will also pursue the integration of scalable model training into DLHub workflows and
work to further decrease model serving latencies.

\section*{Acknowledgments}
This work was supported in part by Laboratory Directed Research and
Development (LDRD) funding from Argonne National Laboratory 
and the RAMSES project, both from the
U.S. Department of Energy under Contract 
DE-AC02-06CH11357. We thank Amazon Web Services for research credits 
and Argonne for computing resources.

\bibliographystyle{IEEEtran}
\bibliography{refs}

\begin{thebibliography}{10}
\providecommand{\url}[1]{#1}
\csname url@samestyle\endcsname
\providecommand{\newblock}{\relax}
\providecommand{\bibinfo}[2]{#2}
\providecommand{\BIBentrySTDinterwordspacing}{\spaceskip=0pt\relax}
\providecommand{\BIBentryALTinterwordstretchfactor}{4}
\providecommand{\BIBentryALTinterwordspacing}{\spaceskip=\fontdimen2\font plus
\BIBentryALTinterwordstretchfactor\fontdimen3\font minus
  \fontdimen4\font\relax}
\providecommand{\BIBforeignlanguage}[2]{{%
\expandafter\ifx\csname l@#1\endcsname\relax
\typeout{** WARNING: IEEEtran.bst: No hyphenation pattern has been}%
\typeout{** loaded for the language `#1'. Using the pattern for}%
\typeout{** the default language instead.}%
\else
\language=\csname l@#1\endcsname
\fi
#2}}
\providecommand{\BIBdecl}{\relax}
\BIBdecl

\bibitem{balaprakash2016automomml}
P.~Balaprakash \emph{et~al.}, ``{AutoMOMML}: Automatic multi-objective modeling
  with machine learning,'' in \emph{International Conference on High
  Performance Computing}.\hskip 1em plus 0.5em minus 0.4em\relax Springer,
  2016, pp. 219--239.

\bibitem{GoogleAutoML}
``{Google Cloud AutoML},'' \url{https://cloud.google.com/automl/}. Accessed
  October 14, 2018.

\bibitem{sagemaker}
``{Amazon SageMaker},''
  \url{https://docs.aws.amazon.com/sagemaker/latest/dg/whatis.html}. Accessed
  October 14, 2018.

\bibitem{avsec2018kipoi}
Z.~Avsec \emph{et~al.}, ``Kipoi: Accelerating the community exchange and reuse
  of predictive models for genomics,'' \emph{bioRxiv}, vol. 10.1101/375345,
  2018.

\bibitem{crankshaw2017clipper}
D.~Crankshaw \emph{et~al.}, ``Clipper: A low-latency online prediction serving
  system,'' in \emph{14th {USENIX} Symposium on Networked Systems Design and
  Implementation ({NSDI})}, 2017, pp. 613--627.

\bibitem{miao2017towards}
H.~Miao \emph{et~al.}, ``Towards unified data and lifecycle management for deep
  learning,'' in \emph{33rd Intl Conf.\ on Data Engineering}.\hskip 1em plus
  0.5em minus 0.4em\relax IEEE, 2017, pp. 571--582.

\bibitem{jia2014caffe}
Y.~Jia \emph{et~al.}, ``Caffe: Convolutional architecture for fast feature
  embedding,'' in \emph{22nd ACM Intl. Conf.\ on Multimedia}, 2014, pp.
  675--678.

\bibitem{abadi2016tensorflow}
M.~Abadi \emph{et~al.}, ``Tensor{F}low: A system for large-scale machine
  learning,'' in \emph{OSDI}, vol.~16, 2016, pp. 265--283.

\bibitem{chollet2017deep}
F.~Chollet, \emph{Deep Learning with Python}.\hskip 1em plus 0.5em minus
  0.4em\relax Manning Publications, 2017.

\bibitem{pedregosa2011scikit}
F.~Pedregosa \emph{et~al.}, ``Scikit-learn: Machine learning in {P}ython,''
  \emph{Journal of Machine Learning Research}, vol.~12, no. Oct, pp.
  2825--2830, 2011.

\bibitem{chard2016globus}
K.~Chard \emph{et~al.}, ``Globus: Recent enhancements and future plans,'' in
  \emph{XSEDE16 Conference on Diversity, Big Data, and Science at Scale}.\hskip
  1em plus 0.5em minus 0.4em\relax ACM, 2016, p.~27.

\bibitem{tfserving}
C.~Olston \emph{et~al.}, ``{TensorFlow-Serving}: Flexible, high-performance
  {ML} serving,'' in \emph{31st Conf.\ on Neural Information Processing
  Systems}, 2017.

\bibitem{gundersen2017state}
O.~E. Gundersen \emph{et~al.}, ``State of the art: Reproducibility in
  artificial intelligence,'' in \emph{30th AAAI Conf.\ on Artificial
  Intelligence}, 2017.

\bibitem{gossett2018aflowml}
E.~Gossett \emph{et~al.}, ``{AFLOW-ML: A RESTful API for machine-learning
  predictions of materials properties},'' \emph{Computational Materials
  Science}, vol. 152, pp. 134--145, 2018.

\bibitem{zhang2018ocpmdm}
Q.~Zhang \emph{et~al.}, ``{OCPMDM: Online computation platform for materials
  data mining},'' \emph{Chemometrics and Intelligent Laboratory Systems}, vol.
  177, no. November 2017, pp. 26--34, 2018.

\bibitem{agrawal2018fatiguepredictor}
A.~Agrawal \emph{et~al.}, ``An online tool for predicting fatigue strength of
  steel alloys based on ensemble data mining,'' \emph{International Journal of
  Fatigue}, vol. 113, pp. 389--400, 2018.

\bibitem{baker16reproducibility}
M.~Baker, ``1,500 scientists lift the lid on reproducibility,'' \emph{Nature},
  vol. 533, pp. 452--454, 2016.

\bibitem{morin2012shining}
A.~Morin \emph{et~al.}, ``Shining light into black boxes,'' \emph{Science},
  vol. 336, no. 6078, pp. 159--160, 2012.

\bibitem{brinkman18wholetale}
A.~Brinckman \emph{et~al.}, ``Computing environments for reproducibility:
  Capturing the ``{Whole Tale}",'' \emph{Future Generation Computer Sys.},
  2018.

\bibitem{stodden16reproducibility}
V.~Stodden \emph{et~al.}, ``Enhancing reproducibility for computational
  methods,'' \emph{Science}, vol. 354, no. 6317, pp. 1240--1241, 2016.

\bibitem{towns2014xsede}
J.~Towns \emph{et~al.}, ``{XSEDE}: Accelerating scientific discovery,''
  \emph{Computing in Science \& Engineering}, vol.~16, no.~5, pp. 62--74, 2014.

\bibitem{pordes2007open}
R.~Pordes \emph{et~al.}, ``The {Open Science Grid},'' in \emph{Journal of
  Physics: Conference Series}, vol.~78, no.~1.\hskip 1em plus 0.5em minus
  0.4em\relax IOP Publishing, 2007, p. 012057.

\bibitem{blaiszik2016materials}
B.~Blaiszik \emph{et~al.}, ``The {Materials Data Facility}: Data services to
  advance materials science research,'' \emph{JOM}, vol.~68, no.~8, pp.
  2045--2052, 2016.

\bibitem{jain2013commentary}
A.~Jain \emph{et~al.}, ``The {Materials Project}: A materials genome approach
  to accelerating materials innovation,'' \emph{APL Materials}, vol.~1, no.~1,
  p. 011002, 2013.

\bibitem{jupyter}
T.~Kluyver \emph{et~al.}, ``Jupyter {N}otebooks--a publishing format for
  reproducible computational workflows.'' in \emph{ELPUB}, 2016, pp. 87--90.

\bibitem{CaffeModelZoo}
``{Caffe Model Zoo},'' \url{http://caffe.berkeleyvision.org/model_zoo.html}.
  Accessed October 14, 2018.

\bibitem{ModelHubai}
``{ModelHub},'' \url{http://modelhub.ai/}. Accessed October 14, 2018.

\bibitem{olson2018system}
R.~S. Olson \emph{et~al.}, ``A system for accessible artificial intelligence,''
  in \emph{Genetic Programming Theory and Practice XV}.\hskip 1em plus 0.5em
  minus 0.4em\relax Springer, 2018, pp. 121--134.

\bibitem{olston2017tensorflow}
C.~Olston \emph{et~al.}, ``Tensorflow-{S}erving: Flexible, high-performance
  {ML} serving,'' \emph{arXiv preprint arXiv:1712.06139}, 2017.

\bibitem{Kubeflow}
``{Kubeflow},'' \url{https://www.kubeflow.org/}. Accessed April 1, 2018.

\bibitem{ananthakrishnan18publication}
R.~Ananthakrishnan \emph{et~al.}, ``Globus platform services for data
  publication,'' in \emph{Practice and Experience on Advanced Research
  Computing}.\hskip 1em plus 0.5em minus 0.4em\relax ACM, 2018, pp. 14:1--14:7.

\bibitem{hintjens2013zeromq}
P.~Hintjens, \emph{{ZeroMQ}: Messaging for Many Applications}.\hskip 1em plus
  0.5em minus 0.4em\relax O'Reilly Media, Inc., 2013.

\bibitem{parsl}
Y.~Babuji \emph{et~al.}, ``Parsl: Scalable parallel scripting in {P}ython,'' in
  \emph{10th International Workshop on Science Gateways}, 2018.

\bibitem{GlobusAuth}
R.~Anathankrishnan \emph{et~al.}, ``Globus {A}uth: A research identity and
  access management platform,'' in \emph{16th Intl Conf.\ on e-Science}, 2016.

\bibitem{szegedy2016rethinking}
C.~Szegedy \emph{et~al.}, ``Rethinking the {I}nception architecture for
  computer vision,'' in \emph{IEEE Conference on Computer Vision and Pattern
  Recognition}, 2016, pp. 2818--2826.

\bibitem{krizhevsky2009learning}
A.~Krizhevsky, ``Learning multiple layers of features from tiny images,''
  Citeseer, Tech. Rep., 2009.

\bibitem{ong2013python}
S.~P. Ong \emph{et~al.}, ``{Python Materials Genomics} (pymatgen): A robust,
  open-source {P}ython library for materials analysis,'' \emph{Computational
  Materials Science}, vol.~68, pp. 314--319, 2013.

\bibitem{ward2018matminer}
L.~Ward \emph{et~al.}, ``Matminer: An open source toolkit for materials data
  mining,'' \emph{Computational Materials Science}, vol. 152, pp. 60--69, 2018.

\bibitem{ward2016magpie}
------, ``A general-purpose machine learning framework for predicting
  properties of inorganic materials,'' \emph{npj Computational Materials},
  vol.~2, p. 16028, 2016.

\bibitem{Kirklin2015}
S.~Kirklin \emph{et~al.}, ``{The Open Quantum Materials Database (OQMD)}:
  Assessing the accuracy of {DFT} formation energies,'' \emph{npj Computational
  Materials}, vol.~1, p. 15010, 2015.

\bibitem{michie1968memo}
D.~Michie, ```memo' functions and machine learning,'' \emph{Nature}, vol. 218,
  no. 5136, p.~19, 1968.

\bibitem{wozniak2017candle}
J.~M. Wozniak \emph{et~al.}, ``{CANDLE/Supervisor}: A workflow framework for
  machine learning applied to cancer research,'' in \emph{Computational
  Approaches for Cancer Workshop}, 2017.

\bibitem{chard2018high}
R.~Chard \emph{et~al.}, ``High-throughput neuroanatomy and trigger-action
  programming: A case study in research automation,'' in \emph{1st
  International Workshop on Autonomous Infrastructure for Science}.\hskip 1em
  plus 0.5em minus 0.4em\relax ACM, 2018, p.~1.

\end{thebibliography}

\end{document}